\documentclass[letterpaper, 10 pt, conference]{ieeeconf}  

\IEEEoverridecommandlockouts                              

\usepackage{amsmath} 
\usepackage{amssymb}  

\usepackage{url}
\usepackage{graphicx}
\usepackage{xcolor}
\usepackage[color=lightgray!20,textsize=scriptsize]{todonotes}
\usepackage{tikz}
\usepackage{amsmath}
\usepackage{multirow}
\usepackage{colortbl}
\usepackage{xcolor}
\usepackage{array}
\usepackage{wrapfig}
\usepackage{listings}
\usepackage{soul}
\usepackage{tabularx}
\usepackage{bm}
\usepackage{float}
\usepackage{booktabs}
\usepackage[backend=bibtex,natbib=true,sorting=none,giveninits=true,maxbibnames=99,url=false,doi=false]{biblatex}
\usepackage[skip=3pt, font={small}]{caption}
\usepackage[colorlinks=false]{hyperref}
\definecolor{lightgreen}{RGB}{220, 237, 191}
\definecolor{lightblue}{RGB}{221, 235, 247}
\definecolor{firebrick}{HTML}{B22222}   

\renewcommand{\baselinestretch}{1.0} 
\normalsize
\title{\LARGE \bf
InterReal: A Unified Physics-Based Imitation Framework for Learning Human–Object Interaction Skills
}

\author{Dayang Liang$^{\dag 1}$ \quad Yuhang Lin$^{\dag 2}$ \quad Xinzhe Liu$^{3}$ \quad Jiyuan Shi$^{4}$\quad Yunlong Liu$^{\star 1}$ \quad Chenjia Bai$^{\star 4}$
\thanks{$ ^\star$Corresponding Author, $^{\dag}$Equal Contribution}
\thanks{$^{1}$Xiamen University, $^{2}$Zhejiang University, $^{3}$ShanghaiTech University, $^{4}$Institute of Artificial Intelligence (TeleAI), China Telecom}
}
\makeatletter
\let\@oldmaketitle\@maketitle
\renewcommand{\@maketitle}{\@oldmaketitle
  \centering
  \setcounter{figure}{0}%
  \begin{minipage}{\linewidth}
    \includegraphics[width=\textwidth]{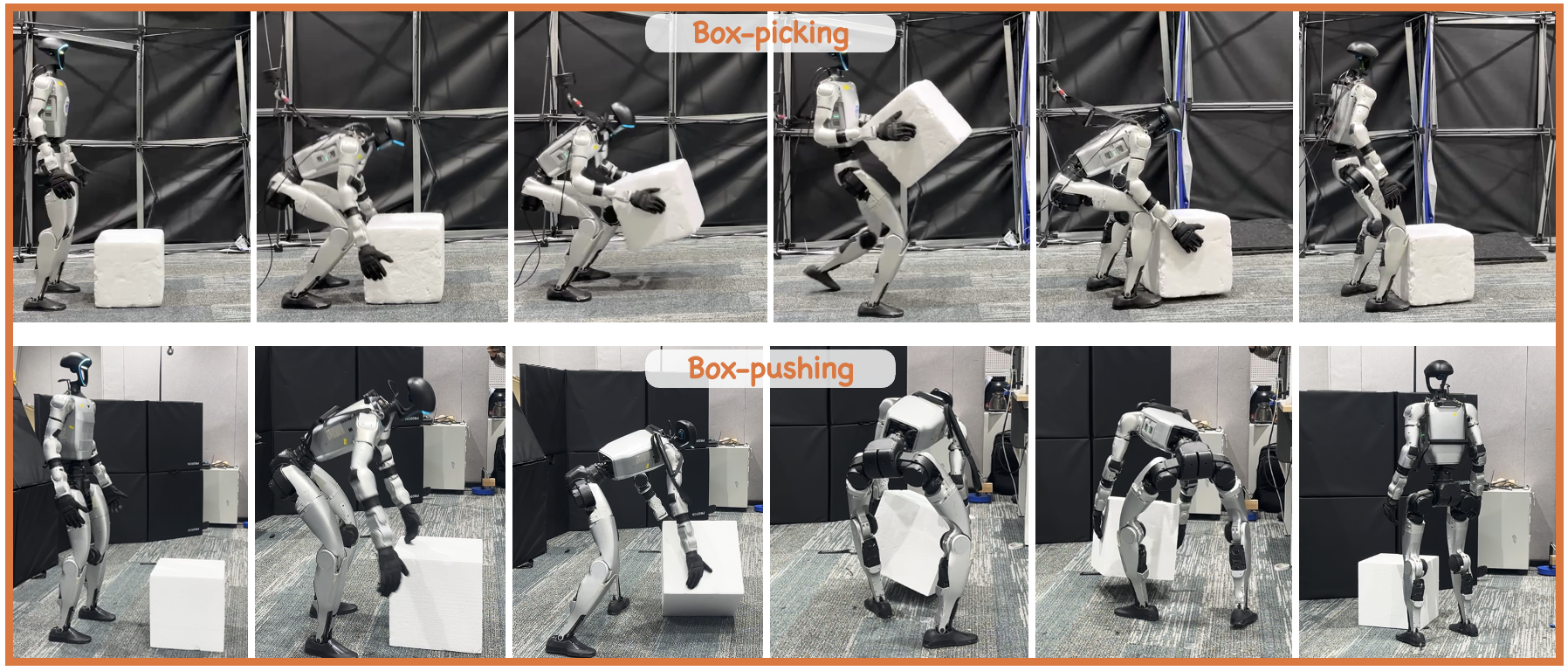}
    {\captionsetup{hypcap=false}%
    \captionof{figure}{\label{fig6}
    Two groups of live photos from a real-world deployment of the challenging interaction tasks.
\textcolor{orange}{\textbf{Top group}}: The robot visually perceives the box's posture, simultaneously picking, walking, and putting the high-density box down.
\textcolor{orange}{\textbf{Bottom group}}: The robot needs to bend slightly and continuously push the box forward. During the interaction, the policy can be adjusted in real-time based on unfavorable box postures to ensure HOI task completion.}}
  \end{minipage}
  \vspace{-10pt}
}
\makeatother

\begin{document}
\maketitle


\thispagestyle{empty}
\pagestyle{empty}

\begin{abstract}

Interaction is one of the core abilities of humanoid robots. However, most existing frameworks focus on non-interactive whole-body control, which limits their practical applicability. In this work, we develop InterReal, a unified physics-based imitation learning framework for \underline{Real}-world human–object \underline{Inter}action (HOI) control. InterReal enables humanoid robots to track HOI reference motions, facilitating the learning of fine-grained interactive skills and their deployment in real-world settings. Within this framework, we first introduce a HOI motion data augmentation scheme with hand–object contact constraints, and utilize the augmented motions to improve policy stability under object perturbations. Second, we propose an automatic reward learner to address the challenge of large-scale reward shaping. A meta-policy guided by critical tracking error metrics explores and allocates reward signals to the low-level reinforcement learning objective, which enables more effective learning of interactive policies. Experiments on HOI tasks of box-picking and box-pushing demonstrate that InterReal achieves the best tracking accuracy and the highest task success rate compared to recent baselines. Furthermore, we validate the framework on the real-world robot Unitree G1, which demonstrates its practical effectiveness and robustness beyond simulation.

\end{abstract}

\section{Introduction}

Deep reinforcement learning (DRL) with motion-imitation has achieved remarkable progress on humanoid robots~\cite{peng2018deepmimic}. These approaches have allowed real-world robots to realize a variety of highly dynamic and complex whole-body control policies, such as walking~\cite{radosavovic2024real}, jumping~\cite{asap}, and dancing~\cite{liao2025beyondmimic}. However, due to the lack of interactive ability, these skills cannot accurately accomplish complex human-object interaction (HOI) tasks, which limits their applicability in real-world scenarios such as industrial application~\cite{raees2024explainable}.

In the real world, developing precise and feedback-driven HOI policies for humanoid robots via motion-imitation learning remains a significant challenge~\cite{carfi2021hand}. Early studies on the policies can be traced back to the field of physical animation. For example, works such as InterMimic~\cite{xu2025intermimic} and CooHOI~\cite{gao2024coohoi} explored DRL to enable virtual characters to track human–object reference motions and execute complex, physical interactive behaviors. However, they typically ignored the full constraints of real-world physics, often relying on animation-level physics settings and lacking fine-grained modeling of contact mechanisms, which makes them difficult to deploy directly to real-world robotic systems. In robotics, existing interaction solutions primarily rely on upper-body teleoperation~\cite{ben2025homie}, where human operators control the robot’s interactive behavior. While effective, such methods constrain the autonomy of humanoid robots. More recently, advanced whole-body teleoperation controllers such as CLONE~\cite{li2025clone}, TWIST~\cite{ze2025twist}, and OpenWBT~\cite{zhang2025unleashing} have been proposed. They leverage large-scale mocap datasets to train fundamental controllers with the ability of general motion tracking. Although such methods can promptly generate interactive behaviors, they remain limited by issues of stability and control precision, as well as the lack of consideration of hand-object contact issues.

To address these limitations, we aim to provide a unified HOI training framework to support real-world deployment and interactive feedback. Unlike whole-body control policies or teleoperation controllers, the framework simultaneously tracks human–object motion trajectories and integrates both observations into DRL training, aiming to achieve accurate and smooth closed-loop interaction tasks. Nevertheless, achieving this HOI goal presents substantial challenges. First, our empirical investigation indicates that during real-world deployment, sensor disturbances in human and object trajectories, particularly relative position perturbations between them, can easily drive the learned policy out of distribution or even cause collapse. Although prior work has introduced domain randomization (DR), motion blending, and other generalization tricks to alleviate sim-to-real gaps~\cite{tobin2017domain}, research specifically targeting interactive generalization remains scarce. Furthermore, reward design has long been a fundamental bottleneck in DRL for humanoid robots with complex settings, as balancing multiple reward signals with heterogeneous purposes while approximating an optimal reward function is a notoriously difficult task~\cite{schulman2017proximal}.

In this paper, we develop InterReal with two key components: HOI motion augmentations and an automatic reward learner, designed to mitigate HOI disturbances and the hard trade-offs in reward design. First, in the motion augmentation stage, we use inverse kinematics (IK) to ensure consistency of hand-object contact details with the anchor motion, while generating multiple motions by altering object positions. Training with these augmented motions significantly improves the generalization ability of interactive policies. Second, our automatic reward learner is motivated by the insight: we observe that DRL tasks based motion-imitation have a clear high-level objective, i.e., minimizing several key position tracking errors. This inspires us to design a meta-DRL policy guided by these error metrics to dynamically balance the large-scale reward terms within the HOI policy. During training, the meta-policy continuously explores optimal reward weights as changes in the errors, effectively improving motion tracking performance.

Experimentally, we evaluate InterReal against recent methods on HOI motions of {box-picking} and box-pushing task. InterReal achieves the lowest tracking error on key metrics, such as DOF angles and object positions, across both tasks. Furthermore, our framework achieved the highest task success rate on both tasks. Additionally, our ablation study shows that the automatic reward mechanism enables learning more effective policies than fixed heuristic rewards. Finally, we reproduced the tasks on the real-world Unitree G1 robot with real-time object posture feedback, demonstrating the effectiveness and robustness of the interactive framework.

\section{Related Work}
\textbf{Physics-based Humanoid Locomotion.}~Previous work has explored whole-body control policies based on DRL for humanoid locomotion by designing motion tracking rewards~\cite{zhang2025hilo,dugar2024learning,cheng2024expressive,ji2024exbody2,xie2025kungfubot,he2024learning,he2025hover} and adversarial motion prior rewards~\cite{peng2021amp,li2017mmd,li2023learning,l2024learning,tang2024humanmimic,escontrela2022adversarial,wu2023learning}, allowing proprioceptive movements such as dancing, walking, and running. In addition, teleoperation of humanoids~\cite{shi2025adversarial,lu2025mobile,ding2025jaeger,xue2025unified,zhang2025falcon} combines lower-body DRL policies with upper-body teleoperation, which enables robots to perform more complex interactive tasks. More recently, researchers have trained more fundamental whole-body control systems (e.g. OmniH20~\cite{he2024omnih2o} and TWIST ~\cite{ze2025twist}), which support adaptive responses to action commands and enable complex whole-body teleoperation as well as motion-driven real-time movement tasks. These controllers can further serve as foundation models for generative motion trajectory–based control manner~\cite{tirinzoni2025zero,luo2023universal,yao2022controlvae,zhang2025natural,xue2025leverb,pan2024model}. Nevertheless, these approaches do not explicitly design feedback-driven controllers for interaction and thus struggle to support accurate human–object interaction tasks.

\textbf{Human-Object Interaction.}~Although human-object interaction based on motion-imitation has achieved remarkable outcomes in the field of animation, it remains a challenge to implement human-object interaction strategies based on the real world. Previous work has proposed learning methods~\cite{luo2024omnigrasp,pan2025tokenhsi,xu2025intermimic} such as RMD \cite{liao2024rmd}, TokenHSI~\cite{pan2025tokenhsi}, and InterMimic~\cite{xu2025intermimic} to construct diverse character behaviors in 3D environments and complete HOI tasks for single-agent or multi-agent cooperation. Some of these methods, such as InterMimic, take physical settings into consideration, but they actually only introduce properties such as small mass and friction to meet the requirements of physical animation, which is still an idealized physical world. In this work, our aim is to extend the human-object interaction DRL learners based on motion imitation from animation to humanoid robots based on the real world.

\section{Preliminaries}

\subsection{Problem Formulation} 
The HOI DRL task can be formulated as a Markov Decision Process (MDP) defined by the tuple $( \mathcal{S}, \mathcal{A}, \mathcal{P}, f_t, \gamma )$. Here, $\mathcal{S}$ denotes the perfect state space of the human--object interaction system, $\mathcal A \in \mathbb{R}^{23}$ represents the learnable action space corresponding to the robot’s degrees of freedom (DoFs), $\mathcal{P}$ is the state transition function of the underlying HOI system, $f_t$ denotes the total reward function of the complex interaction system under a given state $s_t$ and action $a_t$, and $\gamma$ is the reward discount factor. The goal of the HOI task is to maximize the expected cumulative discounted reward $\mathbb{E}_{\pi} [ \sum_{t=0}^{T} \gamma^{t} f_t]$. We employ proximal policy optimization (PPO) to optimize the above objective.

In this work, the state $s_t=[s^h_t,s^o_t,s^{\rm int}_t,p]$ at time $t$ consists of humanoid proprioception features $s^h_t$, object features $s^o_t$, interaction features $s^{int}_t$, and the task phase $p$. Specifically, the humanoid features are defined as $s_t^h=[\boldsymbol{q}_t,\dot{\boldsymbol{q}}_t,\boldsymbol{\omega}_t^{root},\boldsymbol{g}_t,\boldsymbol{a}_{t-1}]$, with joint positions $\boldsymbol{q}_t\in \mathbb{R}^{23}$, joint velocities $\dot{\boldsymbol{q}}_t\in \mathbb{R}^{23}$, root angular velocity $\boldsymbol{\omega}_t^{root}\in \mathbb{R}^{3}$, gravity projection $\boldsymbol{g}_t\in \mathbb{R}^{3}$, and the last action $\boldsymbol{a}_{t-1}\in \mathbb{R}^{23}$; the object features are defined as $s_t^o=[\boldsymbol{q}_t^o,\dot{\boldsymbol{q}}_t^o,\boldsymbol{\vartheta}_t^o]$, with global object position $\boldsymbol{q}_t^o\in \mathbb{R}^{3}$, velocity $\dot{\boldsymbol{q}}_t^o\in \mathbb{R}^{3}$, and orientation $\boldsymbol{\vartheta}_t^o\in \mathbb{R}^{3}$; and the interaction-related features are defined as $s_t^{int}=[\boldsymbol{h}_t^o,\boldsymbol{s}_t^{ig}]$, with contact boolean state  $\boldsymbol{h}_t^o \in \mathbb{R}^{c_1}$ and an interaction graph $\boldsymbol{s}_t^{ig}\in \mathbb{R}^{c_2}$, where $c_1$ and $c_2$ denote the number of selected links on the humanoid and the number of feature points chosen on the object, respectively. Please see Appendix~B for detailed parameters.

\begin{figure*}[th]
\begin{center}
\vspace{-10pt}
\includegraphics[width=0.99\textwidth]{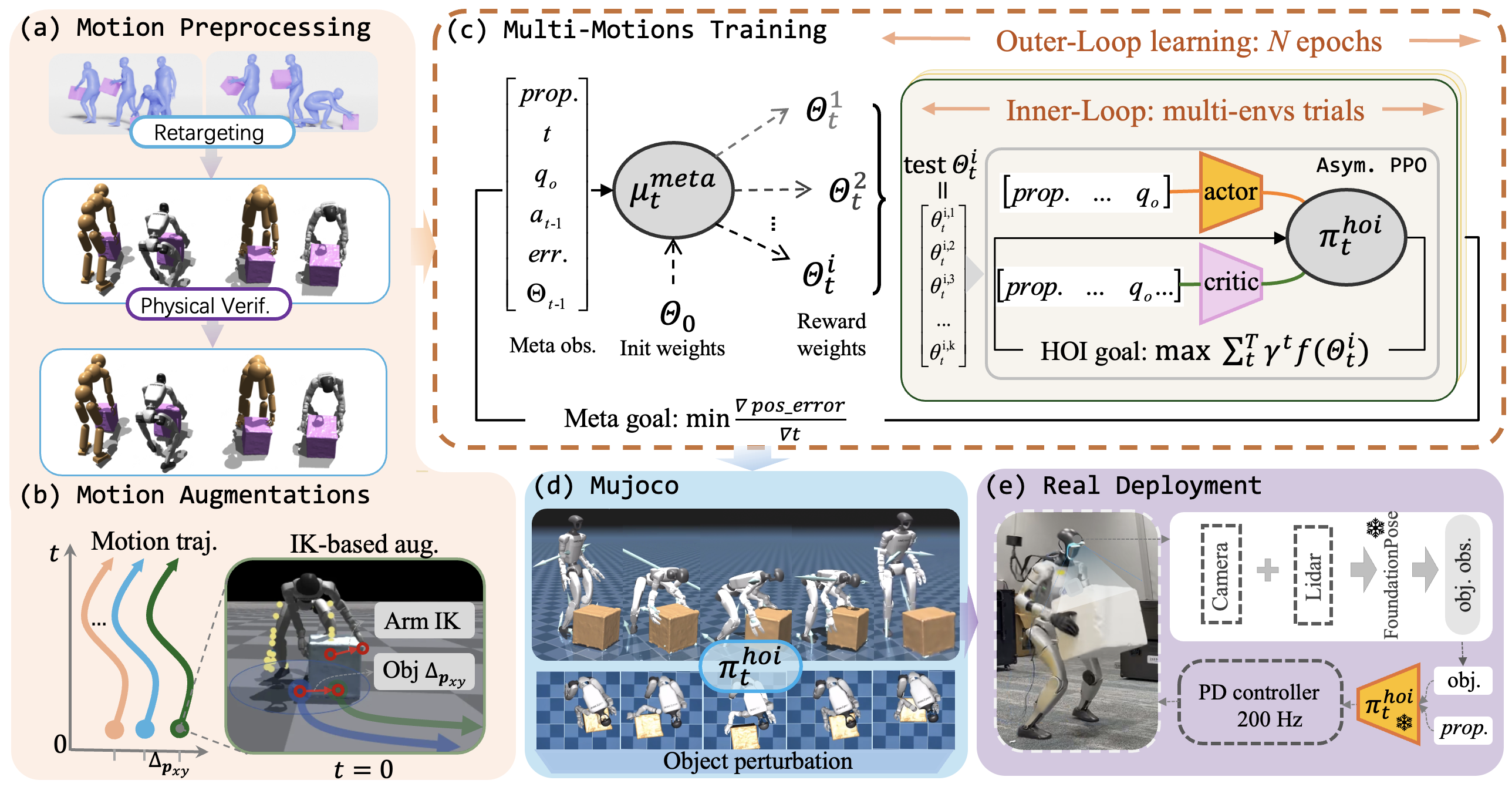}
\end{center}
\caption{
\textbf{Overall framework of InterReal.} InterReal consists of three main components: motion data preprocessing, multi-motion multi-environment learning, and deployment. It enables retargeting HOI motions into trainable motions with G1 robot shapes, achieves accurate motion tracking in complex HOI training settings, and ultimately supports real-world deployment.
\vspace{-5pt}
}
\label{fig1}
\vspace{-5pt}
\end{figure*}

\subsection{Motion-Imitation Learning}

Motion imitation learning has emerged as an important branch of whole-body control learner, with the aim of leveraging DRL to mimic human reference mocap trajectories to ensure fundamental balance, naturalness of movement, and task generalization in humanoid robots. The design of tracking rewards is central to this paradigm. Typically, the similarity between the reference state $s^{ref}$ and the humanoid state $s^{h}$ 
is measured across multiple dimensions such as joint poses, link positions, velocities, and orientations. These similarities are combined into a weighted reward function $f_t$, often in conjunction with phase variables and reference state initialization to mitigate long-term dependency issues and reduce exploration difficulty. To further address the sim-to-real gap, Domain Randomization (DR) over physical properties (e.g., mass and friction) is commonly applied during physics-based training in IsaacGym. Finally, the learned policy outputs the target joint positions of the PD controller, which are then converted into motor torques to reproduce the desired control trajectory. The list of motor PDs used in this work can be found in Appendix~C.

\subsection{Meta Learning}
The goal of meta-learning is to train a meta-learner that can quickly adapt to a new task through mastering prior knowledge about a distribution of tasks. One of its premises is that both the meta-training tasks and the meta-testing tasks are drawn from the same task distribution $p(\mathcal{T})$ and share a common structure. Specifically, the meta-learner can learn a task-specific rule $\Theta'=U(\mathcal{D}^{tr}_{\mathcal{T}},\Theta)$ from the training set $\mathcal{D}^{tr}_{\mathcal{T}}$ and use it to estimate the parameter $\Theta'$ on new tasks, which aims to minimize the lower-learner loss $\mathcal{L}$ with $\Theta'$ and the training parameters $\psi$ in the test set $\mathcal{D}^{test}_{\mathcal{T}}$.
\begin{equation}
\min_{\Theta,\psi}\mathbb{E}_{\mathcal{T}\thicksim p(\mathcal{T})}\left[\mathcal{L}(\mathcal{D}_{\mathcal{T}}^{test},\Theta^{\prime})\right]\mathrm{~s.t.~}\Theta^{\prime}=U(\mathcal{D}_{\mathcal{T}}^{tr},\Theta)
\end{equation}

It is worth noting that the task distribution $p(\mathcal{T})=p(\mathbb{M,\pi})$, where each latent task $\mathcal{T}_i\sim p(\mathcal{T})$ can be described as an MDP $\mathbb{M}_{\mathcal{T}_i}=(\mathcal{S}, \mathcal{A}, \mathcal{P}, f_t, \gamma )$ associated with a lower-level policy $\pi$. Consequently, all tasks share the same structural form but differ in their reward functions. In our problem setting, we introduce the soft actor–critic (SAC) as the meta-learner to learn reward functions for latent subtasks $\{\mathcal{T}_0,\mathcal{T}_1,...,\mathcal{T}_i\}$ in HOI task. This process is referred to as automatic reward learning throughout this paper.

\section{Methodology}
In this section, we present InterReal, a physics-constrained framework for human–object interaction (HOI) control. The overall architecture of InterReal is illustrated in Fig.~\ref{fig1}. We first describe the main pipeline of the InterReal learner, from motion data preprocessing to real-world deployment. We then provide detailed discussions of HOI motion augmentation, followed by the inner-loop HOI policy learning and the outer-loop automatic reward learning process during training.

\subsection{Overall Framework}

\subsubsection{Motion Preprocessing}

As illustrated in Fig.~\ref{fig1} (a) and Fig.~\ref{fig1} (b), preprocessing consists of three steps:  

\textit{HOI Motion Retargeting.} This step aims to retarget mocap data in SMPL format into HOI motions with G1 robot shape parameters. Specifically, for the HOI task, we modify the existing humanoid-centric SMPL motion retargeting approach~\cite{phc} to achieve humanoid retargeting while optimizing the naturalness of contact among the humanoid, object and ground. This particular work will be published later.

\textit{Physical Verification.} Since raw mocap data and retargeted data may contain noise or violate kinematic constraints, directly using them for training can lead to artifacts such as penetration, collisions, invalid references, and low-quality trainings. To address this, we adopt the physics-based HOI tracking approach InterMimic~\cite{xu2025intermimic}, which verifies retargeted motions by enforcing physical constraints in the IsaacGym simulator, thus producing validated HOI motion data.  

\textit{HOI Motion Augmentation.}  As shown in Fig.~\ref{fig1} (b), this step augments a physically validated motion into multiple motions. The resulting multi-motion trajectories correspond to the same task but differ in object positions due to injected offsets, while preserving identical hand–object contact details from the anchor motion. This multi-motion data are then employed in our single-task multi-motion training framework, with the goal of enhancing robustness to object perturbations and improving adaptability in HOI tasks.

\subsubsection{Multi-motion Training} 
Following prior work~\cite{asap,xu2025intermimic}, humanoid skill learning is typically formulated as a DRL task that tracks reference motions. As illustrated in Fig.~\ref{fig1} (c), InterReal introduces two key improvements: i) extending the ASAP framework to track HOI motions, and ii) incorporating a multi-motion HOI tracking learner for augmented motions. The learner consists of two iterative loops: the inner loop learns task-specific HOI policies $\pi^{hoi}$, while the outer loop optimizes a meta-policy $\mu^{meta}$ to learn how to learn $\pi^{hoi}$.

\subsubsection{Deployment} 
As shown in Fig.~\ref{fig1} (d) and Fig.~\ref{fig1} (e), the learned $\pi^{hoi}$ undergoes both sim-to-sim and sim-to-real validation. First, we deploy the policy in the Mujoco simulator~\cite{todorov2012mujoco}, which provides a closer approximation to the real world, to verify its task completion capability and stability, as well as its generalization performance for different object biases. This stage also serves to provide feedback for fine-tuning policy parameters. Subsequently, using the same configuration, we deploy the policy on the real-world humanoid robot Unitree G1. Notably, in real-world deployment, we leverage FoundationPose~\cite{wen2024foundationpose}, a tool capable of performing pose estimation from target meshes, to extract object-related observations.  

\subsection{Motion Augmentation}
To address observation perturbations and enhance generalization in HOI tasks, we introduce a motion augmentation method for HOI motion trajectories. The core idea is to apply an object position offset $\Delta \boldsymbol{p}_{xy}$ along the $XY$-axes of the original HOI motion $\mathcal{M}$ (expressed in the world coordinate system), and then solve the corresponding arm joint positions via inverse kinematics (IK) while maintaining the contact details between the hand links and object.  

Considering the augmentation process of motion $\mathcal{M}^j$, 
since the original motion $\mathcal{M}$ is expressed with the world coordinate system $(\text{w})$, we transform it into the standard pelvis coordinate system $(\text{p})$ for IK solving, so as to obtain new joint positions for both arms. Specifically, we first obtain the object position $\boldsymbol{p}^{(\text{w})}_{obj}(t) \in \mathbb{R}^3$ of the center of mass, the left / right wrist link (end-effectors) positions $\boldsymbol{p}^{(\text{w})}_{L/R}(t) \in \mathbb{R}^3$, and the initial joint angle positions of the left/right arms $\boldsymbol{q}_{\text{arm}}(t) \in \mathbb{R}^{2 \times 7}$ from the original motion at time $t$.  Given the object offset value $\Delta \boldsymbol{p}_{xy}^j = [\Delta x, \Delta y, 0]^{\top}$, we can compute the augmented positions of the object and the left/right end-effectors in the world coordinate system:  
\begin{equation}
{\hat{\boldsymbol{p}}}^{(\text{w})}(t) = \boldsymbol{p}^{(\text{w})}(t) + \mathrm{\Delta}\boldsymbol{p}_{xy}^{j},~{\hat{\boldsymbol{p}}}_{L/R}^{(\text{w})}(t) = \boldsymbol{p}_{L/R}^{(\text{w})}(t) + \mathrm{\Delta}\boldsymbol{p}_{xy}^{j}~
\end{equation}

Secondly, the wrist (end-effector) positions 
$\hat{\boldsymbol{p}}^{(\text{w})}_{L/R}(t)$ are transformed into the coordinate systems of the torso and pelvis through the transformation matrices $T_{(\text{t} \leftarrow \text{w})}$ and $T_{(\text{p} \leftarrow \text{t})}$, respectively:  
\begin{equation}
{\hat{\boldsymbol{p}}}_{L/R}^{(\text{p})}(t) = T_{(\text{p}\leftarrow \text{t})}T_{(\text{t}\leftarrow \text{w})}{\hat{\boldsymbol{p}}}_{L/R}^{(\text{w})}(t)
\end{equation}

Finally, in the pelvis coordinate system, we call the \emph{Ipopt} nonlinear optimizer~\cite{wachter2006implementation} to solve for the new arm joint angles $\boldsymbol{\hat q}_{\text{arm}}(t)$ given the initial $\boldsymbol q_{\text{arm}}(t)$. By iteratively applying this procedure with offsets $\Delta x, \Delta y \in \texttt{linspace}(-\epsilon, \epsilon, \sqrt{c_3})$, we generate the augmented multi-motion $\{\mathcal{M}^j\}^{c_3}_{j=1}$ corresponding to the same HOI task, where the parameters $\epsilon$ and $c_3$ are specified in Appendix~B. 

\subsection{Inner-Loop HOI Task Learning}
The goal of inner-loop learning is to optimize a motion imitation--based humanoid HOI control policy using Proximal Policy Optimization (PPO). Specifically, we design large-scale subreward signals based on reference motion states to guide the $\pi^{hoi}_{\phi}(a_t|s_t)$, then maximizing the weighted sum of $K$ rewards: $\sum_{t}^T \gamma^t f_t(\Theta)$, where weighted reward function $f_t(\Theta) = \sum_{k=1}^K \theta_t^k r^k(t)$ and subreward weights $\Theta = \{\theta^k\}_{k=1}^K$. The subreward terms $\{r^k(t)\}^K_1$ are typically defined as the difference between robot--object states (e.g., joint positions or link positions) and their reference motions, as well as penalty terms on torque or center-of-mass projection (see Appendix~A for details), which encourages the robot--object system to accurately track the reference motion.  In prior work, empirically selecting a large set of reward weights $\Theta$ is tedious and suboptimal. Moreover, this ignores the inherent fact that the reward weights will change as the phase changes, since the emphasis of HOI learning tasks naturally shifts across different motion phases.

We employ the outer-loop meta-policy $\mu^{meta}$ with a learned model $U_{\psi}$ to dynamically construct the reward function during PPO training. $\mu^{meta}$ determines an appropriate set of weights $\Theta$ according to the learning progress of the PPO task. For instance, at the early stage of a box-lifting task, it can recognize that balance is a primary objective and accordingly assign a larger balance-related reward signal to accurately update the value network $V_{\nu}(s_t)$. Based on the adaptive weighted rewards, the PPO objective $\mathcal{L}^{hoi}$ can be simplified as follows:  
\begin{align}
&\mathcal{L}^{hoi}\left( {\phi,\upsilon;\Theta,\mathcal{D}_{t}^{hoi}} \right) = \\\nonumber
&\mathbb{E}_{a_{t},s_{t},s_{t}^{im}\sim\mathcal{D}_{t}^{hoi}}\left\lbrack {-\frac{\pi_{\phi}\left( a_{t} \middle| s_{t}^{im} \right)}{\pi_{\phi^{'}}\left( a_{t} \middle| s_{t}^{im} \right)}\left( {{\sum\limits_{t}^{T}{\sum\limits_{k = 1}^{K}{\theta_{t}^{k}r^{k}(t)}}} - V_{\upsilon}\left( s_{t} \right)} \right)} \right\rbrack
\end{align}
with the gradient updates,
\begin{equation}
\phi\leftarrow\phi + \beta\nabla_{\phi}\mathcal{L}^{hoi}\left( {\phi,\upsilon;~\Theta,\mathcal{D}_{t}^{hoi}} \right),~\upsilon\leftarrow\upsilon + \beta\nabla_{\upsilon}V_{\upsilon}.
\end{equation}
where $s_t^{im}$ and $s_t$ denote imperfect states for the actor network $\pi_{\phi}$, and perfect states $s_t$ for the value network $V_{\upsilon}$, respectively. The replay buffer $\mathcal{D}_t^{\mathrm{hoi}} = \{ s_t, s_t^{\mathrm{im}}, a_t, r(t) \}_{t < t'}^{t'}$ is constructed within the $t$-epoch PPO horizon steps.

\textbf{Interaction-Aware Reward.} Following the work of InterMimic~\cite{xu2025intermimic}, 
we introduce an interaction-graph tracking reward $r_t^{\mathrm{ig}}$ to 
encourage accurate contacts between the robot and the object. 
The reward $r_t^{\mathrm{ig}}$ is derived from the interaction-graph feature $s_t^{\mathrm{ig}}$, which can be regarded as a distance-based representation between the key robot link positions and the object’s key feature points. Then, we define $r_t^{\mathrm{ig}} = \exp \big( - \theta_t^{\mathrm{ig}} \cdot e_t^{\mathrm{ig}} \big)$,
where the tracking error is given by $e_t^{\mathrm{ig}} = \| s_t^{\mathrm{ig}} - s_{\mathrm{ref},t}^{\mathrm{ig}} \|^2$, and $\theta_t^{\mathrm{ig}}$ denotes the interaction-graph reward weight predicted by the outer-loop meta-policy. Detailed initial weights and reward terms can be found in Appendix~A.

\textbf{Asymmetric Actor-Critic.} Considering that some HOI features are difficult to obtain in the real world, we design an asymmetric actor-critic module in the PPO architecture to distill away the privileged features in the actor. Specifically, the critic has access to perfect states consisting of robot proprioception, gravity projection, interaction graph, and object features. In contrast, the actor only receives imperfect states that exclude the interaction graph as well as object velocity and rotation features. It is worth noting that unstable object features such as velocity and rotation significantly amplify the sim-to-real gap and the policy vulnerability. Therefore, the actor states only incorporates the object position features extracted from the FoundationPose detection.

\subsection{Outer-Loop Automatic Reward Learning}
In the outer-loop task, we design a meta-learning structure to learn the optimal MDP with the optimal reward function for the HOI DRL learning problem. Specifically, we define every $N$ inner-loop PPO epochs as a latent subtask $\mathcal{T}_i \sim p(\mathbb{M}, \pi_{\phi})$, represented by $(\mathbb{M}_{\mathcal{T}_i}, \pi_{\phi})$, where $\mathbb{M}_{\mathcal{T}_i} = (\mathcal{S}, \mathcal{A}, \mathcal{P}, f_t(\Theta), \gamma)$. Each subtask shares the same reward function $f_t(\Theta)$ internally. We then aim to learn a parameterized rule $U_{\psi}$ from the training subtask buffer $D_{\mathcal{T}}^{{tr}} = \{ \mathcal{T}_0, \dots, \mathcal{T}_{t / N} \}$, which is able to compute the corresponding $f_t(\Theta)$ across different subtasks, such that the inner-loop PPO policy loss $L^{{hoi}}$ is minimized on the test task set $D_{\mathcal{T}}^{{test}}$:
\begin{align}
    &{\min\limits_{\Theta,\psi}\mathbb{E}_{\mathcal{T}\sim p(\mathcal{T})}}\left\lbrack {\mathcal{L}^{hoi}\left( \phi,v,\Theta';\mathcal{D}^{hoi},\mathcal{D}_{\mathcal{T}}^{test} \right)} \right\rbrack~~\\\nonumber
    &~~~~~~~~~s.t.~~\Theta' = U_{\psi}\left( \Theta;\mathcal{D}_{\mathcal{T}}^{tr} \right)  
\end{align}

Since the outer-loop task can also be formulated as an MDP process, we employ the Soft Actor-Critic (SAC) algorithm to optimize $U_{\psi}$. In this formulation, the inner-loop PPO training tasks serve as the interaction environment. At each step, the agent observes a state $u_t$, executes a weight action $\Theta_t$ for the current PPO subtask, and receives a reward $G_t$ associated with PPO training performance. The simple policy objective of SAC can be formalized as follows,
\begin{equation}
{\min\limits_{\Theta,\mathit{\psi}}{U_{\psi}\left( {\Theta,\mathcal{D}_{\mathcal{T}}^{tr}} \right)}} = {\min\limits_{\Theta,\mathit{\psi}}\mathbb{E}_{\Theta_{t},u_{t}\sim\mathcal{D}_{\mathcal{T}}^{tr}}}\left\lbrack {-log\mu_{\psi}^{meta}\left( \Theta_{t} \middle| u_{t} \right)G_{t}} \right\rbrack
\end{equation}
where $\Theta$ is updated as follows:
\begin{equation}
\left. \Theta'=\Theta^{0}*\sigma(t)\mu_{\psi}^{meta}\left( \Theta_{t} \middle| u_{t} \right),\sigma(t) = clip\left( {1 - \frac{c_{4}}{t},\delta,1.} \right) \right.   
\end{equation}

This update procedure enables exploration based on the initial weights $\Theta^0$ and the learning of reward weights across different PPO subtasks. The initial reward weight $\Theta^0$, the cutoff epoch number $c_4$, and the minimum factor $\delta$ are provided in Appendix~B. 

The design of the outer-loop task reward is crucial. 
We observe that in motion-imitation robot tasks, regardless of the complexity of the learning rewards (e.g., large-scale reward terms), 
their ultimate goal is to minimize the tracking errors between 
the robot trajectory and the reference trajectory. This insight motivates us to select key tracking-error indicators as the learning signal for the meta-policy $\mu^{meta}_\psi$. In other words, we guide the exploration and optimization of $\Theta$ by varying several critical position tracking errors. Consequently, the meta-policy reward $G_t$ is defined as the change in tracking error of the joint position $e_{jp}$, object position $e_{op}$ and link position $e_{lp}$ within a time window $\Delta t$:
\begin{equation}
G_{t} = \nabla_{t}E_{t} = \frac{\mathrm{\Delta}\left( {e_{jp} + e_{op} + e_{lp}} \right)}{\mathrm{\Delta}t}
\end{equation}
where
\begin{align}
   &e_{jp} = \left\| {\boldsymbol{q}_{t} - \boldsymbol{q}_{ref,t}} \right\|^{2},~e_{op} = \left\| {\boldsymbol{q}_{t}^{o} - \boldsymbol{q}_{ref,t}^{o}} \right\|^{2} \\\nonumber
   &~~~~~~~~~~~~~~~e_{lp} = \left\| {\boldsymbol{q}_{t}^{link} - \boldsymbol{q}_{ref,t}^{link}} \right\|^{2} 
\end{align}

\begin{table*}[t]
\caption{Comparison of the best mean tracking accuracy on the {Box-picking} and Box-pushing task. The data records the average of 20 evaluations of the best model on each tracking metric.}
\vspace{-5pt}
\label{table:hoi}
\begin{sc}
\begin{center}
\begin{tabular}{lcccccccc}
\bottomrule
\rowcolor{gray!20}

Box-picking task  &  &  &  &  &  &  &  &\\
\toprule
Tracking Errors & $E_{\mathrm{mpjpe}} \downarrow$  & $E_{\mathrm{mllpe}} \downarrow$  & $E_{\mathrm{mulpe}} \downarrow$ & $E_{\mathrm{m3lpe}} \downarrow$ & $E_{\mathrm{mope}} \downarrow$ & $E_{\mathrm{morae}} \downarrow$ & $E_{\mathrm{mige}}$($e{-7}$)~$\downarrow$ &  $E_{\mathrm{mlrae}} \downarrow$ \\
\midrule
ASAP*    & 0.1634 & 0.0052 & 0.0039  & 0.0034 &   0.0087  & 0.0049  & 19.09 & 0.4926 \\
InterMimic* & 0.1984 & 0.0053 & 0.0043  & 0.0031 &   0.0032  & 0.0049  & 17.08 & 0.4873 \\
InterReal~(ours) & \textbf{0.1076} & \textbf{0.0037} & \textbf{0.0028}  & \textbf{0.0019} &   \textbf{0.0021}  & \textbf{0.0033}  & \textbf{15.70} & \textbf{0.2893} \\
\bottomrule
\rowcolor{gray!20}
Box-pushing task  &  &  &  &  &  &  &  &\\
\midrule
ASAP*    & 0.0783 & \textbf{0.0015} & 0.0032  & 0.0020 &   0.0033  & 0.0014  & 21.41 & 0.1671 \\
InterMimic* & 0.1064 & 0.0018 & 0.0022  & 0.0021 &   0.0017  & 0.0017  & 9.25 & 0.1704 \\
InterReal~(ours) & \textbf{0.0780} & 0.0018 & \textbf{0.0013}  & \textbf{0.0013} &   \textbf{0.0010}  & \textbf{0.0007}  & \textbf{6.04} & \textbf{0.1604} \\
\bottomrule
\end{tabular}
\end{center}
\end{sc}
\end{table*}


\begin{figure*}[t]
\begin{center}
\includegraphics[width=0.999\textwidth]{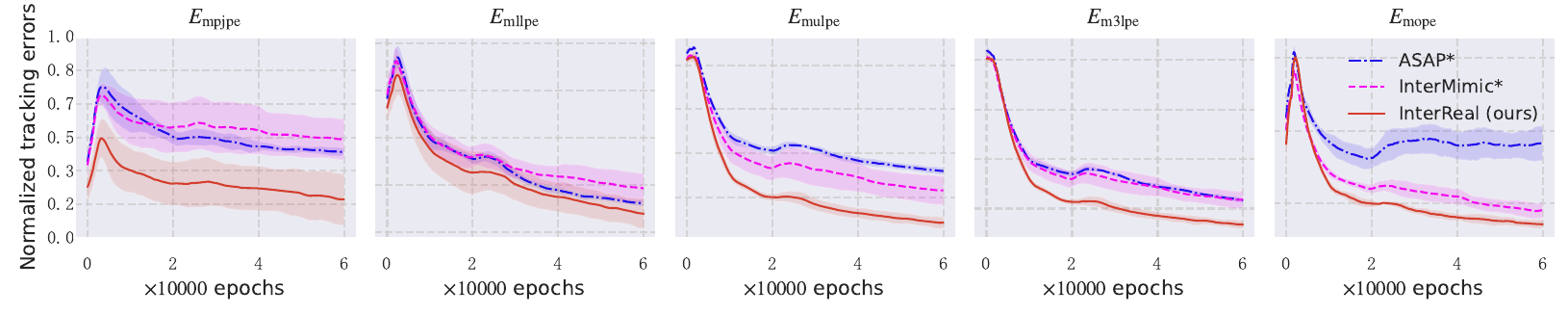}
\end{center}
\caption{
Comparison of tracking accuracy among InterReal and baselines on the {box-picking} task.
}
\label{fig3}
\vspace{-10pt}
\end{figure*}

Finally, we abstract features related to the properties of the inner-loop HOI MDPs as the state representation for the meta-policy. Specifically, the meta-policy state consists of fundamental HOI task features (e.g., $\boldsymbol{q}_t^{{link}}$ and $\boldsymbol{q}_t^{o}$), learning progress, the PPO rewards and actions. These features have been empirically shown to effectively characterize the inner-loop MDPs and support the outer-loop policy learning. Note that in the SAC setting, we define the action entropy as $\mathcal{H} = -\alpha \, \mathbb{E}_{\Theta \sim \mu}[\log \mu^{{meta}}_{\psi}(\Theta|u_t) ]$ with the initial temperature parameter $\alpha = 0.1$, to enhance the exploration of optimal reward function. Overall, the additional computational overhead introduced by the outer-loop learning 
remains minimal, as the outer-loop policy is parameterized in a lightweight manner and trained only once every $N$ inner-loop epochs,

\section{Experiments}
The experiments aim to evaluate the tracking accuracy and the task success rate of the proposed InterReal under training settings that support real-world deployment. Specifically, we first compare InterReal with baseline frameworks on the {box-picking} task (from our mocap dataset) and box-pushing task (from the Omomo dataset \cite{li2023object}). We then conduct ablation studies to verify the effectiveness of the automatic reward component and to analyze the impact of its participation ratio. Furthermore, we visualize how InterReal adapts the reward function to facilitate HOI policy learning. Finally, we deploy the trained HOI policy in the MuJoCo simulator to evaluate its stability and interaction generalization, and to further demonstrate its ability for real-world deployment. Detailed parameters and implementation are provided in \href{https://anonymous.4open.science/r/InterReal\_Appendix/README.md}{supplementary Appendix}.

\textbf{Baselines.}~To our knowledge, most existing open-source HOI motion tracking works focus on the animation domain (e.g., InterMimic \cite{xu2025intermimic}), with very few supporting real-world humanoid robot deployment. Furthermore, recent works targeting the real world primarily concentrate on robot-proprioceptive motion tracking training (e.g., ASAP \cite{asap}). Therefore, for fair comparison, we adapt ASAP and InterMimic into \textbf{ASAP*} and \textbf{InterMimic*}, aligning domain randomization, noise, and key hyperparameters with InterReal. Here, the ASAP* baseline incorporates object tracking in addition to consistent settings compared to the original ASAP. The InterMimic* baseline introduces consistent settings while retaining the basic framework of the original InterMimic. Both variant baselines preserve the fundamental structures of their respective original methods.

\textbf{Evaluation Metrics.}~
Tracking performance is evaluated by the following metrics: Mean Per Joint Position Error ($E_{\mathrm{mpjpe}}$, $rad$), Mean Lower Link Position Error ($E_{\mathrm{mllpe}}$, $m$), Mean Upper Link Position Error ($E_{\mathrm{mulpe}}$, $m$), Mean Three Key Link Position Error ($E_{\mathrm{m3lpe}}$, $m$), Mean Object Position Error ($E_{\mathrm{mope}}$, $m$), Mean Object Rotation Angle Error ($E_{\mathrm{morae}}$, $rad$), Mean Interaction Graph Error ($E_{\mathrm{mige}}$), and Mean Link Rotation Angle Error ($E_{\mathrm{mlrae}}$, $rad$). Additionally, we evaluated the task success rate metric of frameworks through one hundred repeated tests on each HOI task. The calculation of each metric can be found in Appendix~A. 

\begin{figure*}[th]
\begin{center}
\vspace{-10pt}
\includegraphics[width=1\textwidth]{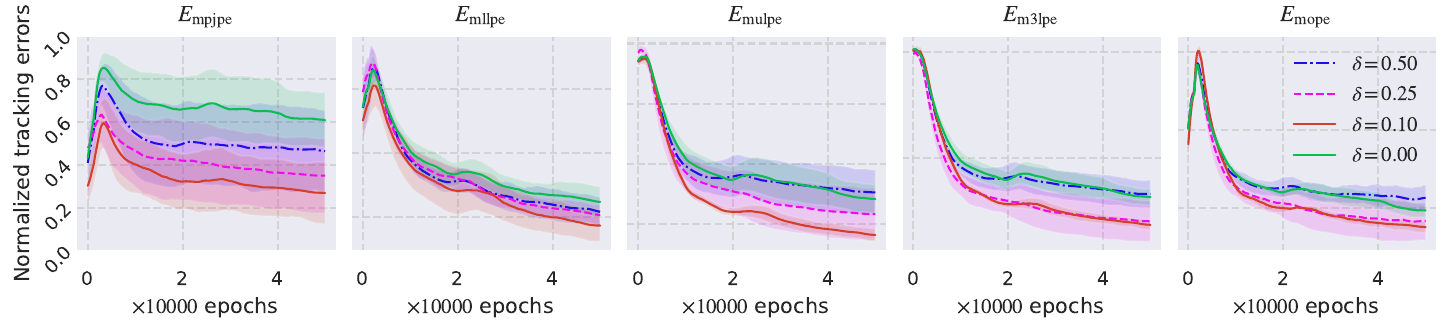}
\end{center}
\vspace{-5pt}
\caption{
Ablation results for the internal coefficient $\delta$ of the meta-learning on the {box-picking} task. 
}
\vspace{-1pt}
\label{fig4}
\end{figure*}

\begin{table}[th]
\caption{Comparison results of task success rate metrics on box-picking and box-pushing tasks.}
\vspace{-5pt}
\label{table:succ}
\begin{sc}
\begin{center}
\begin{tabular}{lcc}
\toprule
Tasks & Box-picking (\%)~$\uparrow$ & Box-pushing (\%)~$\uparrow$  \\
\midrule
ASAP*    & 77.38 & 70.63 \\
InterMimic* & 84.72 & 79.10 \\
InterReal~(ours) & \textbf{96.41} & \textbf{87.45} \\
\bottomrule
\end{tabular}
\end{center}
\vspace{-20pt}
\end{sc}
\end{table}

\subsection{Main Results}
In Table \ref{table:hoi}, we report the best average tracking error for both box-picking and box-pushing HOI tasks. Overall, InterReal achieved the best tracking error on most metrics after training on $5e4$ PPO epochs in both tasks. For example, it achieved the lowest tracking error on both upper link position error $E_{\mathrm{mulpe}}$ and three key link position error $E_{\mathrm{m3lpe}}$. Note that all experiments used uniform domain randomization (DR) settings, detailed DR settings are available in Appendix~B. 

We also show the training curves of normalized error on pick-picking task, as shown in Fig. \ref{fig3}. We observe that InterReal is able to quickly learning and reduce tracking errors, and then achieve the best tracking accuracy overall. However, InterMimic*, which uses a fixed initial reward weight, has significantly higher tracking error, indicating that fixed or handcrafted reward functions are difficult to achieve optimal tracking policies. Finally, ASAP*, due to the lack of interaction graph features and reward design, performs the worst in tracking accuracy, especially for object error. This can be explained by the robot's inability to directly perceive object details through interaction graph features, making it difficult to accurately control the object's position based on the reference motion.

As shown in Table \ref{table:succ}, we evaluated the average success rate of the three methods in each task using the Mujoco simulator, where success was defined as completing the entire HOI task without falling. The data show that the InterReal method achieved the highest success rates than baselines in both tasks, reaching 96.41\% and 87.45\%, respectively, with the box-pushing task being more difficult to succeed than the box-picking task. These empirical results demonstrate that motion enhancement and automatic reward mechanisms can effectively improve the stability and success rate of the framework in handling interactive tasks.

\subsection{Ablation Study}
We show ablation results for the automatic reward learning component and the internal meta-learning coefficient $\delta$, respectively. First, the ablation analysis of the former can be illustrated by the main experimental curves (Fig. \ref{fig3}). The InterMimic* and ASAP* frameworks can essentially be viewed as comparative methods derived by successively ablating the automatic reward component and the interaction graph settings from InterReal. Thus, we can visually observe that both the automatic reward component and the interaction graph significantly enhance the performance of the base framework. Second, $\delta$ acts as a scaling factor for the meta-policy / reward weights, influencing the magnitude of the reward function during the PPO learning process. As shown in Fig. \ref{fig4}, we show the tracking performance of the HOI policy corresponding to $\delta = \{0.0, 0.1, 0.25, 0.50\}$. Visibly, the HOI policy generally achieves the best tracking performance at $\delta=0.1$, with degradation observed at both boundaries and the worst performance at $\delta=0.0$ (i.e., without the automatic reward component). Overall, these analytical results demonstrate the effectiveness of automatic reward learning in guiding the tracking policy.

\subsection{Automatic Reward}
\begin{figure}[th]
\begin{center}
\includegraphics[width=0.5\textwidth]{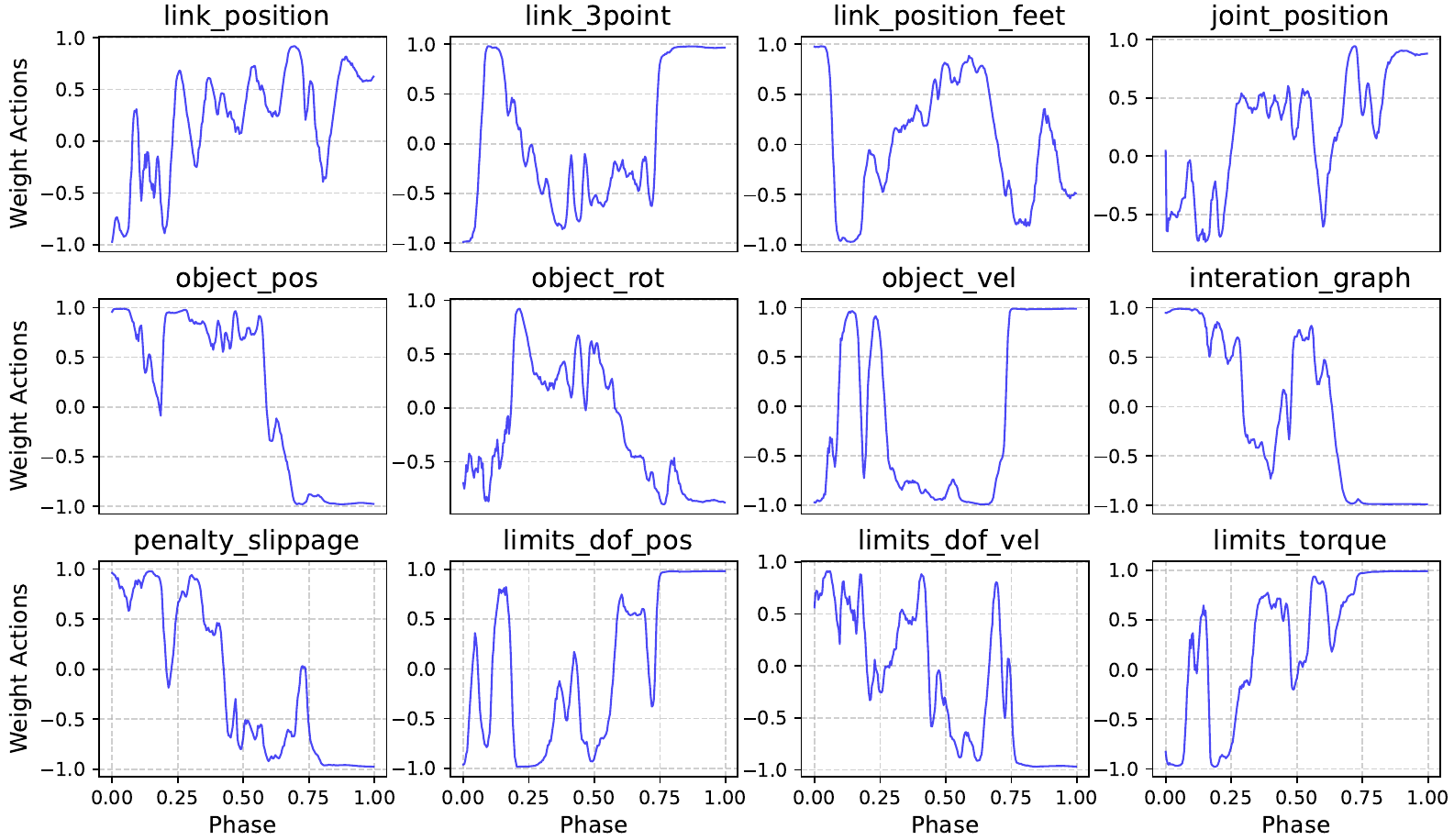}
\end{center}
\vspace{-5pt}
\caption{
Adaptive curves for reward-related weight coefficients.
}
\label{fig5}
\end{figure}

We further analyzed the performance of the proposed meta-policy in the HOI motion imitation task. Fig. \ref{fig5} shows how the reward weights predicted by the meta-policy vary with phase throughout a complete HOI motion phase. As can be seen, the meta-policy adaptively adjusts the relative importance of different reward components, guiding the underlying PPO HOI policy to adopt appropriate control strategies at different subtask stages.

\subsection{Deployment}
As shown in Fig. \ref{fig6}, we demonstrate the capability of the InterReal policy to execute the box task in the real world. Notably, we simulate scenarios such as object position loss, delay, and perturbation during the training phase to ensure policy stability during deployment. Combined with the augmented multi-motion training, the InterReal policy can adjust arm behavior based on different initial object positions and dynamically adjust movement based on the object positions detected by FoundationPose.

\section{CONCLUSION and LIMITATION}
In this paper, we propose a physics-based motion tracking framework, i.e. InterReal for human-object interaction. Motion augmentation and an automatic reward mechanism effectively improve the generalization of the interaction and the efficiency of the policy learning. Compared to existing teleoperation controllers with interactive potential, the InterReal controller can perceive the object's state, enabling more accurate and smooth interactive behaviors. Therefore, this framework has the potential to promote the application of humanoid robots in specific complex interactive tasks.

Despite InterReal's significant interactive advantages, further work is needed to overcome object position perturbations. This issue is primarily caused by the passive object tracking training model and the high variance and latency of object detection in the real world. We will further address these issues in the future.








{
\AtNextBibliography{\scriptsize}
\printbibliography
}


\end{document}